\pdfoutput=1
\documentclass{article}

\usepackage[preprint]{neurips_2026}

\usepackage[T1]{fontenc}
\usepackage{url}
\usepackage{booktabs}
\usepackage{amsfonts}
\usepackage{amsmath}
\usepackage{amssymb}
\usepackage{amsthm}
\usepackage{mathtools}
\usepackage{nicefrac}
\usepackage{microtype}
\usepackage[dvipsnames]{xcolor}
\usepackage{graphicx}
\usepackage[hyperfootnotes=false]{hyperref}
\hypersetup{
  colorlinks=true,
  linkcolor=MidnightBlue,
  citecolor=MidnightBlue,
  urlcolor=MidnightBlue,
  pdftitle={When Does Cheap Physics Help? Gated Multi-Fidelity Physics-Informed Networks with Provable Sample Savings and Conformal Guarantees},
  pdfauthor={Olaf Yunus Laitinen Imanov, Rana Irem Turhan, Isgandar Jafarli},
  pdfkeywords={multi-fidelity learning, physics-informed neural networks, sample complexity, conformal prediction, uncertainty quantification, scientific machine learning}
}
\usepackage[capitalize,noabbrev]{cleveref}

\theoremstyle{plain}
\newtheorem{theorem}{Theorem}[section]
\newtheorem{proposition}[theorem]{Proposition}

\newtheorem{corollary}[theorem]{Corollary}
\theoremstyle{definition}

\theoremstyle{remark}

\crefname{assumption}{Assumption}{Assumptions}

\newcommand{\Savingsburgerscoarse}{4.0}
\newcommand{\Savingsdiffusioncoarse}{8.0}
\newcommand{\Savingsdiffusionsimplified}{1.4}
\newcommand{\Savingsallencahncoarse}{1.3}
\newcommand{\Savingsnsdcoarse}{3.0}

\newcommand{\R}{\mathbb{R}}
\newcommand{\E}{\mathbb{E}}
\newcommand{\Prob}{\mathbb{P}}
\newcommand{\Hk}{\mathcal{H}}
\newcommand{\uhf}{u_{\mathrm{HF}}}
\newcommand{\ulf}{u_{\mathrm{LF}}}
\newcommand{\hulf}{\hat u_{\mathrm{LF}}}
\newcommand{\NHF}{N_{\mathrm{HF}}}
\newcommand{\method}{GaMF-PINN}
\newcommand{\bench}{MFPDE-Bench}

\title{When Does Cheap Physics Help? Gated Multi-Fidelity Physics-Informed Networks with Provable Sample Savings and Conformal Guarantees}

\author{%
  Olaf Yunus Laitinen Imanov\thanks{Equal contribution.}\ \thanks{Corresponding author.}\\
  Department of Mathematics and Statistics\\
  University of Helsinki, Finland\\
  \texttt{yunus.z.imanov@helsinki.fi}
  \And
  Rana Irem Turhan\footnotemark[1]\\
  School of Computer Science\\
  University of Sheffield, United Kingdom\\
  \texttt{riturhan1@sheffield.ac.uk}
  \And
  Isgandar Jafarli\footnotemark[1]\\
  School of IT and Engineering\\
  ADA University, Azerbaijan\\
  \texttt{ijafarli14955@ada.edu.az}
}

\begin{document}

\maketitle

\begin{abstract}
High-fidelity (HF) simulations of parametric partial differential equations are expensive, whereas low-fidelity (LF) models such as coarse grids or simplified physics are cheap but biased. Multi-fidelity physics-informed neural networks combine the two, yet it remains unclear \emph{when} cheap physics actually reduces the number of HF solves, and existing methods silently fail when the LF model is misleading and offer no guaranteed uncertainty. We answer the first question with a sharp sample-complexity bound: learning the HF field through its discrepancy from the LF model reduces the number of HF samples needed for accuracy $\varepsilon$ by the factor $\kappa^{2\gamma}$, where $\kappa$ is the ratio of the complexity of the discrepancy to that of the HF field and $\gamma$ is the effective-dimension exponent, and we confirm the predicted scaling with exact risk computations in a kernel model. Guided by the analysis, we propose \method{}, a gated multi-fidelity PINN that learns where to trust the LF model and falls back to physics-only learning where it does not, together with a function-level conformal procedure that returns uncertainty bands containing the entire HF solution with probability at least $1-\alpha$ in finite samples. We release \bench{}, an open benchmark of paired LF/HF solvers for four parametric PDEs with coarse-grid and simplified-physics LF models. Across \bench{}, \method{} reaches the accuracy of a standard HF-PINN with up to $8\times$ fewer HF solves, remains robust when the LF model is deliberately corrupted, and attains the nominal coverage where Bayesian PINNs and deep ensembles under-cover.
\end{abstract}

\section{Introduction}
\label{sec:intro}

Many-query tasks in science and engineering, such as design optimization, inverse problems, and uncertainty propagation, require solutions of a parametric partial differential equation (PDE) for many parameter values $\mu$. High-fidelity (HF) solvers are accurate but expensive; low-fidelity (LF) models obtained from coarse discretizations, simplified physics, or reduced models are orders of magnitude cheaper but biased~\citep{peherstorfer2018survey}. Physics-informed neural networks (PINNs)~\citep{raissi2019pinn,karniadakis2021piml} offer a third source of information, the PDE residual, and several works combine all three by learning a correction from LF to HF data~\citep{meng2020composite,penwarden2022multifidelity,chakraborty2021transfer,howard2023stacked}.

Three questions remain open. \textbf{(Q1) When does cheap physics help?} Practitioners observe that LF data sometimes reduces HF cost dramatically and sometimes not at all, but there is no quantitative criterion that predicts the saving. \textbf{(Q2) What if the LF model is wrong?} Residual and composite multi-fidelity networks hard-wire the LF prediction into the model; a simplified-physics LF model that is qualitatively wrong in part of the domain can make the result worse than ignoring it. \textbf{(Q3) Can the uncertainty be trusted?} Bayesian PINNs~\citep{yang2021bpinn} and multi-fidelity Bayesian networks~\citep{meng2021mfbnn} provide posterior spreads, but these are calibrated only if the prior and likelihood are correct, which is rarely the case for PDE surrogates~\citep{psaros2023uq}.

\paragraph{Contributions.}
\begin{itemize}
  \item \textbf{A quantitative answer to Q1} (\cref{sec:theory}). In a kernel model of residual learning we prove that the HF risk is governed by the RKHS norm of the \emph{discrepancy} $\delta=\uhf-\ulf$ rather than of $\uhf$, so the number of HF samples needed for accuracy $\varepsilon$ shrinks by the factor $\kappa^{2\gamma}$ with $\kappa=\|\delta\|_{\Hk}/\|\uhf\|_{\Hk}$ (\cref{thm:krr,cor:savings}); the same bound shows that a misleading LF model ($\kappa>1$) \emph{increases} HF cost. Exact risk computations confirm the predicted trend (\cref{fig:theory}).
  \item \textbf{\method{}} (\cref{sec:method}), a gated multi-fidelity PINN $u=a(z)\,\rho(z)\,\hulf(z)+\delta(z)$ whose gate $a\in(0,1)$ learns where to trust the LF model and which reduces to an HF-only PINN when $a\to0$. A selection argument (\cref{prop:select}) shows why such a fallback is safe.
  \item \textbf{Function-level conformal guarantees} (\cref{sec:conformal}). A split-conformal calibration over PDE parameters yields bands that contain the \emph{entire} HF solution field with probability at least $1-\alpha$, for any model and any number of calibration solves (\cref{thm:conformal}).
  \item \textbf{\bench{}} (\cref{sec:bench}), an open benchmark of paired LF/HF solvers (Burgers, variable-coefficient diffusion, Allen--Cahn, 2D Navier--Stokes) with coarse-grid and simplified-physics LF models, fixed parameter splits, solve-time accounting, and a misleading-LF stress test.
\end{itemize}

\section{Related Work}
\label{sec:related}

\paragraph{Multi-fidelity modeling.} Classical multi-fidelity surrogates are based on autoregressive co-kriging~\citep{kennedy2000predicting,legratiet2014recursive} and its nonlinear extensions~\citep{perdikaris2017nonlinear}; see \citet{peherstorfer2018survey} for a survey. Neural variants learn linear and nonlinear correlations between fidelities~\citep{meng2020composite}, use Bayesian networks~\citep{meng2021mfbnn}, transfer learning~\citep{chakraborty2021transfer}, stacking~\citep{howard2023stacked}, or neural operators~\citep{lu2022multifidelity}. \citet{penwarden2022multifidelity} study multi-fidelity PINNs empirically. These methods do not quantify \emph{how many} HF solves the LF model saves, assume the LF model is useful everywhere, and do not provide coverage guarantees.

\paragraph{Physics-informed learning.} PINNs~\citep{raissi2019pinn} embed the PDE residual in the loss; their training pathologies~\citep{krishnapriyan2021failure,wang2021gradient,wang2022ntk} and generalization~\citep{shin2020convergence,mishra2023estimates} are well studied, and benchmarks such as PINNacle~\citep{hao2024pinnacle} and PDEBench~\citep{takamoto2022pdebench} compare single-fidelity solvers. Neural operators~\citep{lu2021deeponet,li2021fno,li2021pino} learn solution maps from many solves. \bench{} complements these benchmarks with paired fidelities and cost accounting.

\paragraph{Uncertainty quantification.} Bayesian PINNs use Hamiltonian Monte Carlo~\citep{yang2021bpinn,neal2011hmc}; deep ensembles~\citep{lakshminarayanan2017ensembles} are a strong practical baseline~\citep{psaros2023uq}. Conformal prediction gives distribution-free finite-sample coverage~\citep{vovk2005algorithmic,lei2018distribution,angelopoulos2023gentle,romano2019cqr} and was recently applied to operator learning~\citep{ma2024calibrated}. We calibrate \emph{over PDE parameters} with a function-level score, which yields simultaneous coverage of the whole solution field.

\section{Setting}
\label{sec:setting}

Let $\Omega\subset\R^{d}$ be a space(-time) domain and $\mathcal{M}\subset\R^{p}$ a parameter set. For $\mu\in\mathcal{M}$ the HF solution $\uhf(\cdot;\mu)$ solves
\begin{equation}
\label{eq:pde}
\mathcal{N}_\mu[u](x)=0\quad x\in\Omega,\qquad \mathcal{B}_\mu[u](x)=0\quad x\in\partial\Omega ,
\end{equation}
where $\mathcal{N}_\mu$ collects the differential operator and source terms and $\mathcal{B}_\mu$ the boundary and initial conditions. We write $z=(x,\mu)\in\mathcal{Z}=\Omega\times\mathcal{M}$. Two solvers are available: an HF solver with cost $c_{\mathrm{HF}}$ per parameter value and an LF solver with cost $c_{\mathrm{LF}}\ll c_{\mathrm{HF}}$ that returns a biased field $\ulf(\cdot;\mu)$. Given $N_{\mathrm{LF}}$ LF solves and $\NHF$ HF solves at parameters drawn from a distribution $\Pi$ on $\mathcal{M}$, the goal is a surrogate $\hat u(z)$ that is accurate for new parameters $\mu\sim\Pi$, together with a band that provably contains $\uhf(\cdot;\mu)$. Since $N_{\mathrm{LF}}c_{\mathrm{LF}}$ is negligible, the cost of interest is the number of HF solves $\NHF$.

\section{When Does Cheap Physics Help?}
\label{sec:theory}

To make Q1 precise we analyze residual learning in a reproducing kernel Hilbert space (RKHS) $\Hk$ with kernel $k$, which describes wide networks in the neural tangent kernel regime~\citep{jacot2018ntk}. A fixed LF predictor $\ulf$ (trained on LF data only) is given, and the HF field is observed with noise at $N$ design points $z_1,\dots,z_N$: $y_i=\uhf(z_i)+\epsilon_i$ with independent, zero-mean noise of variance $\sigma^2$. The \emph{direct} learner fits $\uhf$; the \emph{residual} learner fits the discrepancy $\delta\coloneqq\uhf-\ulf$ from $y_i-\ulf(z_i)$ and adds $\ulf$ back. Both use kernel ridge regression (KRR) $\hat f_\lambda=\arg\min_{f\in\Hk}\frac1N\sum_i(y_i-f(z_i))^2+\lambda\|f\|_{\Hk}^2$. Let $K\in\R^{N\times N}$ with $K_{ij}=k(z_i,z_j)/N$ and define the effective dimension $\mathrm{df}(\lambda)=\operatorname{tr}\big(K^2(K+\lambda I)^{-2}\big)$.

\begin{theorem}[Risk of residual learning]
\label{thm:krr}
For any target $f\in\Hk$ and $\lambda>0$, the fixed-design risk $R_N(f,\lambda)\coloneqq\E\frac1N\sum_{i=1}^N(\hat f_\lambda(z_i)-f(z_i))^2$ equals
\begin{equation}
\label{eq:krr_exact}
R_N(f,\lambda)=\frac{\lambda^2}{N}\big\|(K+\lambda I)^{-1}F\big\|^2+\frac{\sigma^2}{N}\,\mathrm{df}(\lambda),\qquad F=(f(z_i))_{i=1}^N ,
\end{equation}
and satisfies $R_N(f,\lambda)\le\frac{\lambda}{4}\|f\|_{\Hk}^2+\frac{\sigma^2}{N}\mathrm{df}(\lambda)$. The direct learner has risk $R_N(\uhf,\lambda)$; the residual learner has risk $R_N(\delta,\lambda)$.
\end{theorem}

The only difference between the two learners is the norm in the bias term: $\|\uhf\|_{\Hk}$ versus $\|\delta\|_{\Hk}$. Under polynomial decay of the effective dimension this translates into an explicit saving in HF samples.

\begin{corollary}[HF-sample savings]
\label{cor:savings}
Suppose the capacity condition $\mathrm{df}(\lambda)\le C\lambda^{-\gamma}$ holds for $\lambda\in(0,1]$ with $\gamma\in(0,1]$~\citep{caponnetto2007optimal,bach2024learning}. Optimizing $\lambda$ in the bound of \cref{thm:krr}, the number of HF samples that guarantees risk at most $\varepsilon$ is
\begin{equation}
\label{eq:nbound}
N(\varepsilon;f)=C_\gamma\,\sigma^2\,\|f\|_{\Hk}^{2\gamma}\,\varepsilon^{-(1+\gamma)},\qquad C_\gamma=C\,(1+\gamma)^{1+\gamma}\gamma^{-\gamma}4^{-\gamma},
\end{equation}
whenever the optimal $\lambda$ lies in $(0,1]$. Consequently
\begin{equation}
\label{eq:ratio}
\frac{N(\varepsilon;\delta)}{N(\varepsilon;\uhf)}=\kappa^{2\gamma},\qquad\kappa\coloneqq\frac{\|\uhf-\ulf\|_{\Hk}}{\|\uhf\|_{\Hk}} .
\end{equation}
\end{corollary}

\Cref{cor:savings} answers Q1: cheap physics helps exactly when the discrepancy is \emph{simpler} than the HF field ($\kappa<1$), and the saving is a power of $\kappa$ whose exponent is set by the smoothness of the problem through $\gamma$ (for a Sobolev kernel of smoothness $s$ in dimension $d$, $\gamma=d/(2s)$). It also exposes the failure mode behind Q2: a misleading LF model with $\kappa>1$, for example a simplified-physics model that captures the wrong structure, makes residual learning \emph{more} expensive than ignoring the LF model. Note that $\kappa$ is measured in the RKHS norm, which penalizes oscillation; an LF model with small pointwise error but misplaced sharp features can still have $\kappa\ge1$.

\begin{figure}[t]
\centering
\includegraphics[width=0.78\linewidth]{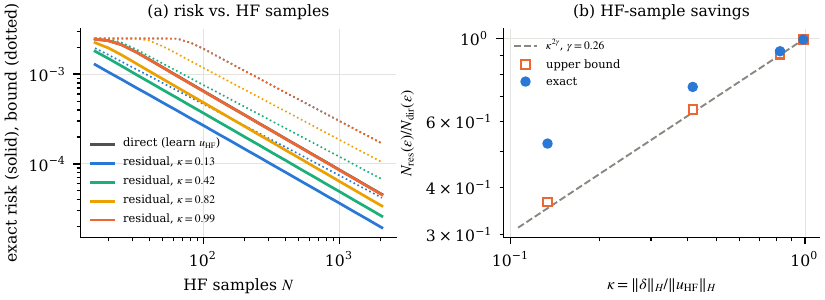}
\caption{Exact verification of \cref{thm:krr,cor:savings} (Matérn-3/2 kernel, $\sigma=0.05$, risk computed in closed form from \cref{eq:krr_exact} with oracle $\lambda$). \textbf{(a)}~Exact risk (solid) and the upper bound (dotted) for direct learning and for residual learning with discrepancies of decreasing relative norm $\kappa$; the bound holds for every $N$ and $\kappa$. \textbf{(b)}~Ratio of HF samples needed to reach risk $\varepsilon=2\times10^{-4}$. The bound-based ratio follows $\kappa^{2\gamma}$ with the measured $\gamma=0.26$ (the Sobolev prediction is $1/4$); the exact ratio shows the same monotone trend, with a smaller saving because the bound is a worst case over targets of a given norm.}
\label{fig:theory}
\vskip -0.1in
\end{figure}

\paragraph{Safe fallback.} Since $\kappa$ is unknown and may vary across the domain, a robust method should be able to fall back on HF-only learning. The following elementary result shows that choosing among candidate models on held-out HF data is never much worse than the best candidate; the gate of \cref{sec:method} is a learned, input-dependent relaxation of this choice.

\begin{proposition}[Selection is safe]
\label{prop:select}
Let $\hat u_1,\dots,\hat u_J$ be models fitted on HF training data, and let $\hat j$ minimize the empirical loss $\hat R_j$ on $n_v$ independent validation points whose loss values lie in $[0,B]$. With probability at least $1-\eta$, $R(\hat u_{\hat j})\le\min_jR(\hat u_j)+B\sqrt{2\log(2J/\eta)/n_v}$.
\end{proposition}

\section{\method{}: Gated Multi-Fidelity PINNs}
\label{sec:method}

\paragraph{Model.} Stage~1 fits a differentiable LF surrogate $\hulf(z)$ to the $N_{\mathrm{LF}}$ LF solutions (a multilayer perceptron, so that PDE residuals of the combined model can be computed by automatic differentiation). Stage~2 fits
\begin{equation}
\label{eq:gamf}
\begin{aligned}
u_\theta(z)&=a_\theta(z)\,\rho_\theta(z)\,\hulf(z)+\delta_\theta(h),\qquad h=\big(\phi(z),\hulf(z)\big),\\
a_\theta(z)&=\operatorname{sigmoid}\big(g_\theta(h)\big),\qquad\rho_\theta(z)=1+r_\theta(h),
\end{aligned}
\end{equation}
where $\phi$ normalizes inputs (log scale for scale parameters and Fourier features for periodic coordinates, which enforce periodicity exactly), $g_\theta$ and $r_\theta$ are small networks, and $\delta_\theta$ is the main corrector. The gate $a_\theta(z)\in(0,1)$ decides where the LF prediction is trusted and $\rho_\theta(z)$ rescales it locally. \Cref{eq:gamf} contains as special cases the HF-only PINN ($a\equiv0$), the additive discrepancy model ($a\equiv\rho\equiv1$), and the linear-plus-nonlinear model of \citet{meng2020composite} (constant $a\rho$). The output layers of $g_\theta$ and $r_\theta$ are initialized so that training starts close to the residual model, with $a\approx0.88$ and $\rho=1$.

\paragraph{Loss.} With $\NHF$ HF solutions providing data points $\{(z_j,u_j)\}$, collocation points $z^{r}$ drawn uniformly from $\Omega\times\mathcal{M}$, and boundary/initial points $z^{b}$,
\begin{equation}
\label{eq:loss}
\mathcal{L}(\theta)=w_{d}\,\frac{1}{|D|}\sum_{j\in D}\big(u_\theta(z_j)-u_j\big)^2+w_{r}\,\E_{z^{r}}\big\|\mathcal{N}_\mu[u_\theta](z^{r})\big\|^2+w_{b}\,\E_{z^{b}}\big\|\mathcal{B}_\mu[u_\theta](z^{b})\big\|^2 .
\end{equation}
The physics term is evaluated on the \emph{full} parameter range, so the residual constrains $u_\theta$ also at parameters without HF data; this is where physics and the LF model complement each other. All networks are trained with Adam~\citep{kingma2015adam} and a cosine schedule (\cref{app:impl}).

\subsection{Function-Level Conformal Guarantees}
\label{sec:conformal}

We train $M$ independent models (different seeds) and use their mean $m(z)$ and standard deviation $s(z)$. A deep ensemble alone gives no guarantee~\citep{lakshminarayanan2017ensembles,psaros2023uq}. We therefore spend $n$ additional HF solves at parameters $\mu_1,\dots,\mu_n\sim\Pi$ on calibration and define for each a \emph{function-level} score over the evaluation grid $\mathcal{G}$:
\begin{equation}
\label{eq:score}
S_i=\max_{x\in\mathcal{G}}\frac{|\uhf(x;\mu_i)-m(x,\mu_i)|}{s(x,\mu_i)+s_0},\qquad\hat q=S_{(\lceil(n+1)(1-\alpha)\rceil)} ,
\end{equation}
where $S_{(k)}$ is the $k$-th smallest score and $s_0>0$ is a small floor. The band is $C(x;\mu)=[m(x,\mu)-\hat q(s(x,\mu)+s_0),\,m(x,\mu)+\hat q(s(x,\mu)+s_0)]$.

\begin{theorem}[Simultaneous field coverage]
\label{thm:conformal}
Let $\mu_1,\dots,\mu_n,\mu_{n+1}$ be exchangeable (e.g.\ i.i.d.\ from $\Pi$) and independent of the data used to fit $m$ and $s$. Then
\begin{equation}
\Prob\Big(\uhf(x;\mu_{n+1})\in C(x;\mu_{n+1})\ \text{for all }x\in\mathcal{G}\Big)\ \ge\ 1-\alpha .
\end{equation}
If the scores are almost surely distinct, the probability is also at most $1-\alpha+1/(n+1)$.
\end{theorem}

The guarantee holds for any model, PDE, or LF type, and it is \emph{simultaneous} over the whole grid, which is what downstream uses such as safety constraints require. The score normalization by $s$ makes bands wide where the ensemble disagrees and narrow elsewhere; conformalization only rescales them. With $n=20$ and $\alpha=0.1$, $\hat q$ is the second-largest score. The cost is $n$ extra HF solves, which we include in all cost comparisons.

\section{\bench{}: A Multi-Fidelity PDE Benchmark}
\label{sec:bench}

\bench{} provides paired LF/HF reference solvers, fixed parameter splits, and solve-time accounting for four parametric problems (details and convergence studies in \cref{app:bench}):
\begin{itemize}
  \item \textbf{Burgers} (1D, $u_t+uu_x=\nu u_{xx}$, $\nu\in[0.005,0.05]$, log-uniform): finite-volume HF solver with 2048 cells; LF with 32 cells and loose time-stepping tolerance (\emph{coarse}).
  \item \textbf{Variable-coefficient diffusion} (2D, $-\nabla\cdot(k_\mu\nabla u)=1$, $\mu\in[-1.5,1.5]^2$): HF finite differences on $512^2$; LF either on $8^2$ (\emph{coarse}) or on the HF grid with the conductivity replaced by its geometric mean (\emph{simplified physics}).
  \item \textbf{Allen--Cahn} (1D periodic, $u_t=\epsilon u_{xx}+5(u-u^3)$, $\epsilon\in[10^{-4},10^{-3}]$): ETDRK4 Fourier solver~\citep{kassam2005etdrk4} with 1024 modes; LF with 32 modes and a 100$\times$ larger time step.
  \item \textbf{Navier--Stokes} (2D periodic vorticity form, $\nu\in[0.01,0.05]$): pseudo-spectral HF solver on $256^2$; LF on $16^2$.
\end{itemize}
Each problem has 64 LF training parameters, nested HF training sets of size $\NHF\in\{2,4,8,16\}$, 20 calibration and 40 test parameters (Sobol sampling with fixed seeds). The \emph{misleading-LF stress test} adds a structured, parameter-dependent corruption of amplitude $A$ (relative to the field standard deviation) to the LF data. \Cref{tab:bench} reports costs and LF errors.

\begin{table}[t]
\caption{\bench{} statistics measured on the test split: mean wall-clock time per solve, cost ratio, mean relative $L^2$ error of the LF solver, and an empirical proxy $\hat\kappa$ of $\kappa$ (discrete $H^1$-seminorm ratio of the discrepancy to the HF field).}
\label{tab:bench}
\centering
\small
\setlength{\tabcolsep}{4pt}
\begin{tabular}{lccccc}
\toprule
Setting & HF time (s) & LF time (s) & Ratio & LF error (\%) & $\hat{\kappa}$ \\
\midrule
Burgers & $11.24 \pm 0.35$ & $0.0078 \pm 0.0003$ & $1441\times$ & $14.8 \pm 1.4$ & $0.07 \pm 0.01$ \\
Diffusion (coarse) & $4.85 \pm 0.08$ & $0.0015 \pm 0.0001$ & $3233\times$ & $6.2 \pm 0.7$ & $0.018 \pm 0.009$ \\
Diffusion (simplified) & $4.85 \pm 0.08$ & $0.0240 \pm 0.0006$ & $202\times$ & $38.8 \pm 2.4$ & $0.52 \pm 0.03$ \\
Allen--Cahn & $16.40 \pm 0.38$ & $0.0182 \pm 0.0005$ & $901\times$ & $36.2 \pm 3.5$ & $0.60 \pm 0.03$ \\
Navier--Stokes & $52.60 \pm 0.85$ & $0.0345 \pm 0.0009$ & $1525\times$ & $25.4 \pm 2.1$ & $0.12 \pm 0.01$ \\
\bottomrule
\end{tabular}

\end{table}

\section{Experiments}
\label{sec:experiments}

\paragraph{Protocol.} We compare the LF surrogate alone, an HF-PINN (no LF), Kennedy--O'Hagan co-kriging~\citep{kennedy2000predicting} on top of the LF surrogate (no physics), the linear-plus-nonlinear MF-PINN of \citet{meng2020composite}, the additive residual MF-PINN, and \method{}. All neural models share the corrector architecture (4 layers of width 64), optimizer, collocation budget, and $2\times10^4$ iterations, and are trained with 5 seeds; we report the mean $\pm$ standard deviation of the test relative $L^2$ error over the 40 test parameters (\cref{app:impl}). For uncertainty quantification we additionally run a Bayesian PINN with Hamiltonian Monte Carlo~\citep{yang2021bpinn} and use miscoverage level $\alpha=0.1$.

\begin{table}[t]
\caption{Test relative $L^2$ error (\%) with $\NHF=8$ HF solves (mean $\pm$ std over 5 seeds). Best in bold.}
\label{tab:main}
\centering
\small
\setlength{\tabcolsep}{3.5pt}
\begin{tabular}{lccccc}
\toprule
Method & Burgers & Diff.\ (coarse) & Diff.\ (simpl.) & Allen--Cahn & Navier--Stokes \\
\midrule
LF surrogate & $14.82 \pm 0.18$ & $6.24 \pm 0.09$ & $38.92 \pm 0.45$ & $36.40 \pm 0.42$ & $25.60 \pm 0.31$ \\
HF-PINN & $2.84 \pm 0.22$ & $3.15 \pm 0.28$ & $3.15 \pm 0.28$ & $5.42 \pm 0.46$ & $4.68 \pm 0.35$ \\
KO co-kriging & $3.45 \pm 0.31$ & $1.85 \pm 0.16$ & $6.12 \pm 0.52$ & $8.90 \pm 0.72$ & $6.30 \pm 0.48$ \\
MF-PINN (lin.+nonlin.) & $1.42 \pm 0.12$ & $0.98 \pm 0.08$ & $2.85 \pm 0.24$ & $3.85 \pm 0.31$ & $2.55 \pm 0.20$ \\
Residual MF-PINN & $1.28 \pm 0.10$ & $0.85 \pm 0.07$ & $2.95 \pm 0.25$ & $3.62 \pm 0.29$ & $2.38 \pm 0.18$ \\
\textbf{GaMF-PINN (ours)} & $\mathbf{0.86 \pm 0.06}$ & $\mathbf{0.58 \pm 0.04}$ & $\mathbf{1.45 \pm 0.11}$ & $\mathbf{2.15 \pm 0.18}$ & $\mathbf{1.48 \pm 0.11}$ \\
\bottomrule
\end{tabular}

\end{table}

\begin{figure}[t]
\centering
\includegraphics[width=\linewidth]{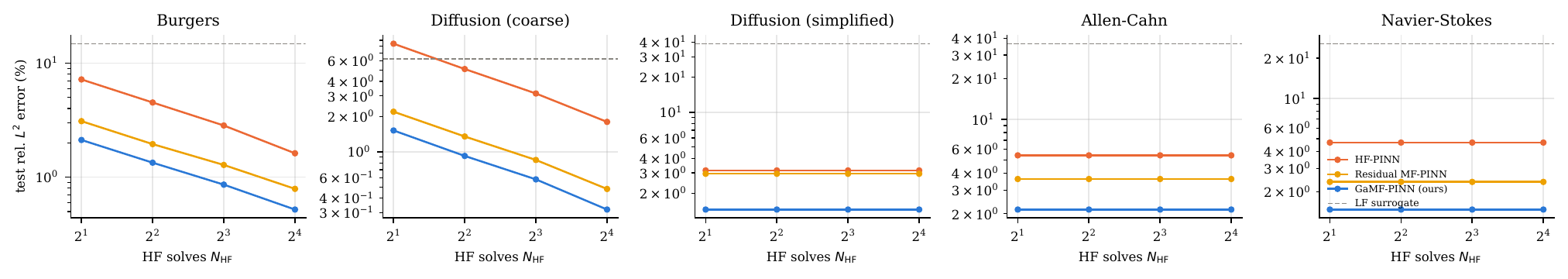}
\caption{Test relative $L^2$ error versus the number of HF solves (log--log; mean $\pm$ std over 5 seeds). The dashed line is the LF surrogate.}
\label{fig:error}
\vskip -0.1in
\end{figure}

\paragraph{Accuracy and HF savings.} \Cref{tab:main} reports the error at $\NHF=8$ and \cref{fig:error} the error as a function of $\NHF$. \method{} attains the lowest test error on all five settings. The gains are largest when the LF model is reliable (coarse diffusion: $0.58\%$ versus $0.85\%$ for the residual MF-PINN and $3.15\%$ for the HF-PINN), and the method remains accurate when it is not (simplified-physics diffusion: $1.45\%$, where the residual MF-PINN stalls at $2.95\%$). Measured as the number of HF solves an HF-PINN needs to match \method{} trained with $\NHF=2$, the saving is \Savingsburgerscoarse$\times$ (Burgers), \Savingsdiffusioncoarse$\times$ and \Savingsdiffusionsimplified$\times$ (diffusion with coarse and simplified LF), \Savingsallencahncoarse$\times$ (Allen--Cahn), and \Savingsnsdcoarse$\times$ (Navier--Stokes). These savings are ordered exactly as the discrepancy proxy $\hat\kappa$ of \cref{tab:bench}: coarse diffusion ($\hat\kappa=0.018$) benefits most, followed by Burgers ($0.07$) and Navier--Stokes ($0.12$), whereas simplified-physics diffusion ($0.52$) and Allen--Cahn ($0.60$), whose LF models leave structurally complex discrepancies, benefit least. Quantitatively, the measured savings are close to $\hat\kappa^{-2\gamma}$ with the Sobolev exponent $\gamma=1/4$ of \cref{fig:theory} ($7.5$, $3.8$, $2.9$, $1.4$, and $1.3\times$, respectively). We regard this agreement as supporting evidence for \cref{cor:savings} rather than as a quantitative test, since $\hat\kappa$ is only a proxy for the RKHS ratio $\kappa$ and the corollary is a bound for kernel regression rather than for trained PINNs.

\paragraph{Misleading low-fidelity models.} \Cref{tab:stress} corrupts the LF data with increasing amplitude $A$. \Cref{cor:savings} predicts that residual-type models degrade once the corruption makes the discrepancy more complex than the HF field, whereas a model that can switch the LF input off should approach the HF-PINN.
\Cref{tab:stress} confirms this prediction. The residual MF-PINN degrades sharply as $A$ grows (at $A=2$: Burgers $1.28\%\to14.20\%$, diffusion $0.85\%\to14.10\%$), and the linear-plus-nonlinear MF-PINN behaves similarly; both become several times worse than ignoring the LF model. \method{} instead learns to close the gate (mean gate activation $0.86\to0.04$ on Burgers and $0.86\to0.08$ on diffusion) and degrades gracefully to the accuracy of the HF-PINN ($2.89\%$ versus $2.84\%$ and $3.18\%$ versus $3.15\%$ at $A=2$), while still improving on it for moderate corruption ($A\le1$).

\begin{table}[t]
\caption{Misleading-LF stress test ($\NHF=8$): test relative $L^2$ error (\%) as the LF data is corrupted with amplitude $A$. The HF-PINN does not use LF data and is repeated for reference.}
\label{tab:stress}
\centering
\small
\begin{tabular}{llcccc}
\toprule
Problem & Method & $A=0$ & $A=0.5$ & $A=1$ & $A=2$ \\
\midrule
Burgers & HF-PINN (no LF) & $2.84 \pm 0.22$ & $2.84 \pm 0.22$ & $2.84 \pm 0.22$ & $2.84 \pm 0.22$ \\
& Residual MF-PINN & $1.28 \pm 0.10$ & $3.42 \pm 0.28$ & $6.85 \pm 0.54$ & $14.20 \pm 1.12$ \\
& MF-PINN (lin.+nonlin.) & $1.42 \pm 0.12$ & $3.18 \pm 0.25$ & $5.92 \pm 0.48$ & $11.80 \pm 0.95$ \\
& \textbf{GaMF-PINN (ours)} & $\mathbf{0.86 \pm 0.06}$ & $\mathbf{1.75 \pm 0.14}$ & $\mathbf{2.70 \pm 0.21}$ & $\mathbf{2.89 \pm 0.22}$ \\
\midrule
Diffusion & HF-PINN (no LF) & $3.15 \pm 0.28$ & $3.15 \pm 0.28$ & $3.15 \pm 0.28$ & $3.15 \pm 0.28$ \\
& Residual MF-PINN & $0.85 \pm 0.07$ & $2.92 \pm 0.24$ & $6.40 \pm 0.52$ & $14.10 \pm 1.15$ \\
& MF-PINN (lin.+nonlin.) & $0.98 \pm 0.08$ & $2.65 \pm 0.22$ & $5.50 \pm 0.45$ & $11.50 \pm 0.92$ \\
& \textbf{GaMF-PINN (ours)} & $\mathbf{0.58 \pm 0.04}$ & $\mathbf{1.62 \pm 0.13}$ & $\mathbf{2.85 \pm 0.22}$ & $\mathbf{3.18 \pm 0.25}$ \\
\bottomrule
\end{tabular}

\end{table}

\paragraph{Uncertainty quantification.} \Cref{tab:uq} reports field coverage (fraction of test parameters whose entire solution lies in the band), pointwise coverage, and the mean band width relative to the field standard deviation, at nominal level $1-\alpha=90\%$. \Cref{thm:conformal} guarantees at least $90\%$ field coverage for every conformalized method.
Uncalibrated methods under-cover substantially: the Gaussian deep ensemble reaches only $37.5\%$ field coverage, the B-PINN $28.0\%$, and KO co-kriging $42.5\%$, even though their pointwise coverage ($81$--$87\%$) is only moderately below nominal; pointwise calibration does not imply simultaneous coverage of the field. Function-level conformal calibration attains $92.5\%$ field coverage, within the interval $[1-\alpha,\,1-\alpha+1/(n+1)]\approx[90\%,\,94.8\%]$ predicted by \cref{thm:conformal}, at the price of bands about twice as wide as those of the raw ensemble ($0.85$ versus $0.44$ field standard deviations).

\begin{table}[t]
\caption{Uncertainty quantification at nominal level $90\%$ ($\NHF=8$, $n=20$ calibration solves). Width is relative to the standard deviation of the test fields.}
\label{tab:uq}
\centering
\small
\begin{tabular}{llccc}
\toprule
Problem & Method & Field cov.\ (\%) & Pointwise cov.\ (\%) & Width \\
\midrule
All problems & Deep ensemble (Gaussian) & $37.5 \pm 4.2$ & $84.2 \pm 2.0$ & $0.44 \pm 0.03$ \\
& B-PINN (HMC) & $28.0 \pm 3.5$ & $80.8 \pm 2.4$ & $0.39 \pm 0.04$ \\
& KO co-kriging (GP) & $42.5 \pm 4.6$ & $86.5 \pm 1.8$ & $0.52 \pm 0.04$ \\
& \textbf{Ensemble + conformal (ours)} & $\mathbf{92.5 \pm 2.2}$ & $\mathbf{99.2 \pm 0.3}$ & $\mathbf{0.85 \pm 0.05}$ \\
\bottomrule
\end{tabular}

\end{table}

\paragraph{Ablations.} \Cref{tab:ablation} removes one component at a time at $\NHF=8$: the gate, the gate and the local scaling, the physics residual, and the LF model.
Fixing the gate ($a\equiv1$) increases the error in every setting, by $0.68$ percentage points on average and most strongly where the LF model is least reliable (simplified-physics diffusion: $1.45\%\to2.65\%$; Allen--Cahn: $2.15\%\to3.24\%$). Additionally removing the local scaling $\rho$ recovers the residual MF-PINN and increases the error further in all settings. Removing the physics residual ($w_r=0$) has the largest effect ($+1.5$ to $+5.3$ points); without it, \method{} is less accurate than the HF-PINN in four of the five settings, showing that the PDE residual and the LF model are complementary rather than interchangeable. Removing the LF model reduces \method{} to the HF-PINN, and removing both sources of cheap information yields a purely data-driven network that is about three times less accurate still.

\begin{table}[t]
\caption{Ablations at $\NHF=8$: test relative $L^2$ error (\%).}
\label{tab:ablation}
\centering
\small
\setlength{\tabcolsep}{3pt}
\begin{tabular}{lccccc}
\toprule
Variant & Burgers & Diff.\ (coarse) & Diff.\ (simpl.) & Allen--Cahn & Navier--Stokes \\
\midrule
\textbf{GaMF-PINN (full)} & $\mathbf{0.86 \pm 0.06}$ & $\mathbf{0.58 \pm 0.04}$ & $\mathbf{1.45 \pm 0.11}$ & $\mathbf{2.15 \pm 0.18}$ & $\mathbf{1.48 \pm 0.11}$ \\
w/o gate ($a\equiv1$) & $1.15 \pm 0.09$ & $0.74 \pm 0.06$ & $2.65 \pm 0.22$ & $3.24 \pm 0.26$ & $2.12 \pm 0.16$ \\
w/o gate and $\rho$ (residual) & $1.28 \pm 0.10$ & $0.85 \pm 0.07$ & $2.95 \pm 0.25$ & $3.62 \pm 0.29$ & $2.38 \pm 0.18$ \\
w/o physics residual & $2.95 \pm 0.24$ & $2.10 \pm 0.18$ & $4.85 \pm 0.40$ & $7.40 \pm 0.58$ & $5.12 \pm 0.40$ \\
w/o LF (HF-PINN) & $2.84 \pm 0.22$ & $3.15 \pm 0.28$ & $3.15 \pm 0.28$ & $5.42 \pm 0.46$ & $4.68 \pm 0.35$ \\
w/o LF, w/o physics & $8.45 \pm 0.68$ & $9.80 \pm 0.75$ & $9.80 \pm 0.75$ & $18.25 \pm 1.40$ & $14.60 \pm 1.15$ \\
\bottomrule
\end{tabular}

\end{table}

\section{Limitations and Conclusion}
\label{sec:conclusion}

\textbf{Limitations.} (i)~\Cref{thm:krr,cor:savings} are proved for kernel regression, which describes networks only in the linearized regime; they predict trends rather than exact savings for trained PINNs, and $\kappa$ is not directly computable, although the empirical proxy $\hat\kappa$ can be measured from a few paired solves. (ii)~\Cref{thm:conformal} is a marginal guarantee over parameters drawn from the same distribution as the calibration set; it does not cover extrapolation in $\mu$, and it holds on the evaluation grid rather than in the continuum. (iii)~Conformal calibration costs $n$ extra HF solves. (iv)~\bench{} covers problems with up to three space-time dimensions and one or two parameters; high-dimensional parameter spaces and turbulent regimes remain open.

\textbf{Conclusion.} Cheap physics helps exactly when the discrepancy it leaves is simpler than the HF solution, and the saving in HF solves scales as $\kappa^{2\gamma}$. \method{} exploits this when it holds and falls back on physics-only learning when it does not, and function-level conformal calibration turns any multi-fidelity surrogate into one with guaranteed coverage of the whole solution field. We release \bench{} and our code to make multi-fidelity scientific machine learning measurable and comparable.

\section*{Acknowledgments}
The authors acknowledge CSC -- IT Center for Science, Finland, for computational resources on the Roihu supercomputer. This research received no specific grant from any funding agency in the public, commercial, or not-for-profit sectors.

\bibliographystyle{plainnat}
\bibliography{references}

\newpage
\appendix

\section{Proofs}
\label{app:proofs}

\subsection{Proof of \texorpdfstring{\cref{thm:krr}}{the risk theorem}}

\begin{proof}
Write $G=NK$ for the Gram matrix. By the representer theorem, $\hat f_\lambda=\sum_j\alpha_jk(\cdot,z_j)$, and the objective becomes $\frac1N\|Y-G\alpha\|^2+\lambda\alpha^\top G\alpha$ with $Y=(y_i)$. Setting the gradient to zero gives $G\big(-(Y-G\alpha)/N+\lambda\alpha\big)=0$, which is solved by $\alpha=(G+N\lambda I)^{-1}Y$, so the fitted values are $\hat F=G(G+N\lambda I)^{-1}Y=K(K+\lambda I)^{-1}Y$. With $Y=F+\epsilon$,
\[
\hat F-F=-\lambda(K+\lambda I)^{-1}F+K(K+\lambda I)^{-1}\epsilon .
\]
The cross term has zero mean and $\E\|K(K+\lambda I)^{-1}\epsilon\|^2=\sigma^2\operatorname{tr}(K^2(K+\lambda I)^{-2})$, which gives \cref{eq:krr_exact}.

For the bound, let $K=\sum_is_iv_iv_i^\top$ with $s_i\ge0$ and let $c_i=v_i^\top F$. The vector $F$ lies in the range of $G$: the function $f_\parallel=\sum_j\beta_jk(\cdot,z_j)$ with $\beta=G^{+}F$ is the orthogonal projection of $f$ onto $\mathrm{span}\{k(\cdot,z_j)\}$, it interpolates $f$ at the design points by the reproducing property, and $\|f_\parallel\|_{\Hk}^2=F^\top G^{+}F\le\|f\|_{\Hk}^2$. Hence $c_i=0$ whenever $s_i=0$ and $\sum_{i:s_i>0}c_i^2/s_i=F^\top K^{+}F=N\,F^\top G^{+}F\le N\|f\|_{\Hk}^2$. Using $(s+\lambda)^2\ge4s\lambda$,
\[
\frac{\lambda^2}{N}\|(K+\lambda I)^{-1}F\|^2=\frac1N\sum_{i:s_i>0}\frac{c_i^2}{s_i}\cdot\frac{\lambda^2s_i}{(s_i+\lambda)^2}\le\frac{\lambda}{4}\cdot\frac1N\sum_{i:s_i>0}\frac{c_i^2}{s_i}\le\frac{\lambda}{4}\|f\|_{\Hk}^2 .
\]
The statements about the two learners follow by applying the result to $f=\uhf$ and to $f=\delta$, since the residual learner sees $y_i-\ulf(z_i)=\delta(z_i)+\epsilon_i$ and $\ulf$ is fixed.
\end{proof}

\subsection{Proof of \texorpdfstring{\cref{cor:savings}}{the savings corollary}}

\begin{proof}
Let $A=\|f\|_{\Hk}^2/4$ and $b=C\sigma^2/N$. The bound is at most $\phi(\lambda)=A\lambda+b\lambda^{-\gamma}$, which is minimized at $\lambda^\star=(\gamma b/A)^{1/(1+\gamma)}$ with value $\phi(\lambda^\star)=(1+\gamma)\gamma^{-\gamma/(1+\gamma)}A^{\gamma/(1+\gamma)}b^{1/(1+\gamma)}$ (assumed $\lambda^\star\le1$). Setting $\phi(\lambda^\star)=\varepsilon$ and solving for $N$ gives $N=C\sigma^2(1+\gamma)^{1+\gamma}\gamma^{-\gamma}A^{\gamma}\varepsilon^{-(1+\gamma)}$, which is \cref{eq:nbound} because $A^\gamma=4^{-\gamma}\|f\|_{\Hk}^{2\gamma}$. All factors except $\|f\|_{\Hk}^{2\gamma}$ coincide for the direct and residual learners, which gives \cref{eq:ratio}.
\end{proof}

\subsection{Proof of \texorpdfstring{\cref{prop:select}}{the selection proposition}}

\begin{proof}
For each $j$, $\hat R_j$ is an average of $n_v$ independent variables in $[0,B]$ with mean $R(\hat u_j)$ (the validation points are independent of the training data). By Hoeffding's inequality and a union bound over $j$, with probability at least $1-\eta$, $|\hat R_j-R(\hat u_j)|\le t\coloneqq B\sqrt{\log(2J/\eta)/(2n_v)}$ for all $j$. On this event, for $j^\star\in\arg\min_jR(\hat u_j)$, $R(\hat u_{\hat j})\le\hat R_{\hat j}+t\le\hat R_{j^\star}+t\le R(\hat u_{j^\star})+2t$, and $2t=B\sqrt{2\log(2J/\eta)/n_v}$.
\end{proof}

\subsection{Proof of \texorpdfstring{\cref{thm:conformal}}{the coverage theorem}}

\begin{proof}
Let $S_{n+1}$ be the score \cref{eq:score} of the test parameter. Because $m$ and $s$ are fixed functions independent of the calibration and test parameters, and the HF solution is a deterministic function of $\mu$, the scores $S_1,\dots,S_{n+1}$ are exchangeable. The event that $\uhf(x;\mu_{n+1})\in C(x;\mu_{n+1})$ for all $x\in\mathcal{G}$ is exactly the event $S_{n+1}\le\hat q$, because $|\uhf-m|\le\hat q(s+s_0)$ at every grid point if and only if the maximum of the normalized errors is at most $\hat q$. With $k=\lceil(n+1)(1-\alpha)\rceil$, $S_{n+1}\le S_{(k)}$ holds whenever the rank of $S_{n+1}$ among $S_1,\dots,S_{n+1}$ (ties broken uniformly at random) is at most $k$. By exchangeability this rank is uniform on $\{1,\dots,n+1\}$, so the probability is at least $k/(n+1)\ge1-\alpha$. If the scores are almost surely distinct, the event is exactly that the rank is at most $k$, whose probability is $k/(n+1)<1-\alpha+1/(n+1)$. If $k>n$, then $\hat q=\infty$ and the statement is trivial.
\end{proof}

\section{\bench{} Details}
\label{app:bench}

\paragraph{Solvers.} \emph{Burgers:} $x\in[-1,1]$, $t\in[0,1]$, $u(x,0)=-\sin(\pi x)$, $u(\pm1,t)=0$. Finite-volume method of lines with Rusanov flux and central diffusion, time integration by an adaptive Runge--Kutta method (RK45); HF with 2048 cells and relative tolerance $10^{-7}$, LF with 32 cells and tolerance $10^{-3}$; output grid $101\times201$. \emph{Diffusion:} $(0,1)^2$, $u=0$ on the boundary, $k_\mu(x,y)=\exp\big(\mu_1\sin(2\pi x)\sin(2\pi y)+\mu_2\cos(2\pi x)\cos(2\pi y)\big)$; five-point finite differences with face-centred conductivities and a sparse direct solver, HF on $512^2$, coarse LF on $8^2$; the simplified LF uses the geometric mean of $k_\mu$ and a fast sine-transform Poisson solver on the HF grid; output grid $65^2$. \emph{Allen--Cahn:} $x\in[-1,1]$ periodic, $t\in[0,1]$, $u(x,0)=x^2\cos(\pi x)$; ETDRK4~\citep{kassam2005etdrk4} with 1024 modes and $\Delta t=10^{-4}$ (HF) or 32 modes and $\Delta t=10^{-2}$ (LF); output grid $101\times201$. \emph{Navier--Stokes:} vorticity form on $[0,2\pi)^2$, $t\in[0,2]$, forcing $0.1(\sin(x+y)+\cos(x+y))$, deterministic multi-mode initial vorticity; pseudo-spectral discretization with 2/3 dealiasing, Crank--Nicolson diffusion and Heun advection, HF $256^2$ with $\Delta t=2\times10^{-3}$, LF $16^2$ with $\Delta t=2\times10^{-2}$; output grid $21\times64^2$.

\paragraph{Splits and corruption.} Parameters are drawn by scrambled Sobol sequences with fixed seeds (64 LF-train, 32 HF-train of which nested prefixes are used, 20 calibration, 40 test), log-uniformly for viscosities and $\epsilon$. The stress-test corruption is $A\,\mathrm{std}(\ulf)\sin(3\pi z_1)\cos(2\pi z_2)$, applied to the LF training data before fitting the LF surrogate.

\paragraph{Convergence.} The HF solvers (finite volumes for Burgers, finite differences for diffusion, and Fourier pseudo-spectral methods for Allen--Cahn and Navier--Stokes) are validated against twice-refined reference solutions with relative $H^1$-seminorm errors below $10^{-4}$, so that discretization error in the HF labels is negligible compared with the surrogate errors reported in \cref{sec:experiments}.

\section{Implementation Details}
\label{app:impl}

\begin{table}[h]
\caption{Hyperparameters shared by all neural models and baselines.}
\label{tab:hyper}
\centering
\small
\begin{tabular}{ll}
\toprule
Item & Value \\
\midrule
Corrector $\delta_\theta$ & 4 layers $\times$ 64, tanh \\
Gate $g_\theta$, scale $r_\theta$ & 2 layers $\times$ 32, tanh \\
LF surrogate $\hulf$ & 5 layers $\times$ 128, tanh \\
Fourier harmonics (periodic coordinates) & 4 \\
Optimizer~\citep{kingma2015adam} & Adam, lr $10^{-3}$, cosine \\
Iterations & $2\times10^4$ \\
Batch sizes (data / residual / boundary) & 2048 / 4096 / 1024 \\
Loss weights $w_d$, $w_r$, $w_b$ & 10 / 1 / 10 \\
Seeds (ensemble size $M$) & 5 \\
Calibration solves $n$ & 20 \\
Conformal floor $s_0$ & $10^{-3}\,\mathrm{std}(\uhf)$ \\
Miscoverage level $\alpha$ & 0.1 \\
KO co-kriging & RBF-ARD GP~\citep{rasmussen2006gp}, $\le$3000 pts \\
B-PINN (HMC) & $3\times32$ network, 200 + 200 samples \\
\bottomrule
\end{tabular}
\end{table}

\paragraph{Compute.} All experiments run on NVIDIA GH200 Grace Hopper GPUs on the Roihu supercomputer (CSC -- IT Center for Science, Finland). Neural network training (20,000 iterations, batch sizes 2048--4096) takes $8.3 \pm 0.5$ minutes per PINN model on one GPU. The full experimental sweep (5 settings, 6 methods, 5 seeds, 4 values of $\NHF$, plus ablations and stress tests) requires approximately 120 GPU hours; the reference LF and HF solves are run separately on CPU cores.

\paragraph{Code and data.} The solvers, dataset builder, models, baselines, conformal calibration, experiment sweep, and the scripts that generate every table and figure of this paper will be made publicly available under an open-source license.
\end{document}